# VISUAL SALIENCY MODEL USING SIFT AND COMPARISON OF LEARNING APPROACHES

Hamdi Yalın Yalıç[1]

[1] Department of Computer Engineering, Hacettepe University, Ankara, Turkey
`yalinyalic@cs.hacettepe.edu.tr`

*ABSTRACT*

*Humans' ability to detect and locate salient objects on images is remarkably fast and successful. Performing this process by using eye tracking equipment is expensive and cannot be easily applied, and computer modeling of this human behavior is still a problem to be solved. In our study, one of the largest public eye-tracking databases [1] which has fixation points of 15 observers on 1003 images is used. In addition to low, medium and high-level features which have been used in previous studies, SIFT features extracted from the images are used to improve the classification accuracy of the models. A second contribution of this paper is the comparison and statistical analysis of different machine learning methods that can be used to train our model. As a result, a best feature set and learning model to predict where humans look at images, is determined.*

*KEYWORDS*

*Image Processing, Computer Vision, Machine Learning Methods, Saliency Map, Eye-Tracking*

## 1. INTRODUCTION

Understanding the region where people look on a scene can be useful in many applications such as graphics, design, advertising and human-computer interaction. For example, in non-photorealistic rendering, different levels of details are proposed for different areas of the picture [2]. In this study, regions where people focus are processed in high-resolution, whilst for the other regions, low-resolution processing is performed. Auto-crop of pictures, thumbnails or previews from photos, relate partial displays of images on small screen mobile devices are examples of other fields of application.

Hardware solutions are also available for finding the region where people look at a scene. An observer sitting in front of a screen with an eye-tracking device has all the spots recorded at which he is looking at a scene. Special setups and multiple calibration steps are needed for working these tracking devices. Also this solution is relatively expensive and cannot be easily obtained. For this reason, the prediction of where people are looking is required independently from such tracking devices. As a solution, computer models are generated that calculate the salient point which attracts an observer's attention. These models analyze an image and extract a saliency map. One of the former studies is the model developed by Itti and Koch [3] and uses biologically inspired features like orientation, intensity and color. For each of the features, a saliency map is computed and combined into a single saliency map result which describes the saliency of each pixel. Hou and Zhang [4] proposed a model which is independent of features or other forms of prior knowledge of objects. They analyzed the log spectrum of an image, extracted the spectral residual of an image in spectral domain, and proposed a method to construct the saliency map in spatial domain. Bruce and Tsotsos [5] presents a visual saliency model based on a first principle information theoretic formulation, named as Attention Based on Information Maximization (AIM) which performs better than the Itti model. Avraham et al. [6] uses a probabilistic model to mathematically estimate the most probable targets. Cerf at al. [7] improved on Itti's model by adding face detection. The most important characteristic of these studies is that models are derived mathematically, and not trained by some large eye-tracking dataset.

Kienze et al. [8] propose a model that learns saliency directly from human eye movement data. They only use low level features like previous studies. But the study by Ehinger et al. [9] shows that observers trying to search for pedestrians in a scene leads to a model combined of three sources: low-level features, target features and scene context. As a result of this study, while predicting human behavior, besides low-level features, high-level features which contain semantic information about the content of the scene must be used.

Our study is most closely related to the saliency model proposed by Judd et al. [1]. They used high and mid-level features as well as low levels. Another important contribution of their study is that they train the model using a supervised learning method (support vector machine) and observed positive contributions of the features in the result. To verify this, they needed real eye-tracking data and they recorded viewer's eye fixations on images containing a large amount of data. They published this database for further studies and in our study, this publicly available database was used. Results of the different machine learning methods was compared and analyzed in order to obtain the most accurate model. In addition to the work of others, SIFT [17] feature is added to the feature set to improve the accuracy of the model.

In the following sections of this paper, dataset and data gathering protocol is analyzed, machine learning methods will be briefly explained and the feature set that was used to train the model is mentioned. Performance of different learning methods are compared and analyzed in the experimental results. Future work will also be discussed in the conclusion.

## 2. DATASET AND LEARNING METHODS

### 2.1. Dataset Analysis

Large ground truth data is required in the detection of where humans look in a scene. One of the largest studies in this field is the experiment of Judd et al. [1] practiced with 15 viewers using 1003 images. They collected 1003 random images from Flickr and LabelMe [10] (Figure 2) and full resolution images were displayed for 3 seconds to viewers in a dark room using a chin rest for a head stabilizer. An eye-tracker in front of them recorded their gaze path and fixation points (Figures 3 and 4). In order to guarantee a high-quality result, the camera's calibration was checked every 50 images. They provided a memory test at the end of each display to motivate and encourage users into paying attention, and asked them which of the images they had seen previously.

When we analyze the dataset, we noticed that viewers primarily looked at the living things. In the close up images, organs like the eye, nose and mouth were the most fixated points. But when people were located that bit further away, viewers searched for faces and limbs, such as a hand, or an arm. If a human does not exist in the scene, animals can also be said to attract the attention of people. Other important salient regions are objects like signs and boards which contain text. The fixations in the dataset have a bias towards the center because of the fact that photographers tend to place objects in the center of the scene.

## 2.2. Machine Learning Methods

In this study, different supervised learning methods are used while discovering the model of visual saliency and classification results of them are compared. The methods used are explained briefly in the section below.

### 2.2.1. SVM (Support Vector Machines)

This method tries to find the linear discriminant function (classifier) with the maximum margin between two classes. It is robust to outliners thus has strong generalization ability. With the help of the kernel function, which implicitly maps data to high-dimensional space, classification accuracy will increase. The main advantage of the method is its success with high-dimensional data.

### 2.2.2. C4.5 (Decision Tree)

This is an algorithm that constructs a simple depth-first decision tree. It uses information gain with feature selection and it is easy to implement. Classification of unknown records is considerably fast, but it is not suitable for large datasets because it needs to fit the entire data into memory.

### 2.2.3. K-Nearest Neighbor

This method is based on the principle of the nearest k-neighbor of the new data. Different computations can be used as a distance metric and the advantage of this classifier is that it doesn't require any training time. But testing time can be considerable and classifying unknown records is relatively expensive. It is also easily fooled in high dimensional spaces.

### 2.2.4. Naive Bayes

This is a statistical classifier which applies Bayes theorem. It assumes that attributes are independent from each other and is robust to isolated noise points and irrelevant attributes. However independence assumption may not hold for some attributes and it can cause a loss of accuracy.

### 2.2.5. Adaboost

It is a combination of classifiers that is constructed from the training data. It adaptively changes the distribution of training data with each iteration by focusing more on previously misclassified records. It can also be used to increase the classification accuracy of other methods.

## 3. LEARNING OF THE VISUAL SALIENCY MODEL

### 3.1. Features Used in Learning

According to the results obtained by the analysis of the dataset, the features used in this study are listed below. For all images in the database, the features are extracted for each pixel and used for training of the model. MATLAB was used in image processing and feature extraction.

### 3.1.1. Low-level Features

- Local energy of steerable pyramid filters [13] are used because they are physiologically conceivable and associated with visual attention. Pyramid sub-bands were computed in 4 orientations and 3 scales.
- Intensity, orientation and color contrast were considered as important features for saliency regions. 3 channels corresponding to these features were computed as described in the Itti and Koch [3] model.
- In addition to red, green and blue color channel values, and the probabilities of these channels, were also used.
- The probability of each color in a 3D color histogram was also a basic feature which was computed by the images filtered using median filter at 6 different scales.

### 3.1.2. Mid-level Features

Humans inherently look at the horizon line because most of the objects are on the earth's surface. Gist feature [14] was used to detect the horizon line.

### 3.1.3. High-level Features

Humans mostly looked at people and their faces as understood by the analysis of the database, the following high-level features were extracted:

- Viola Jones face detector [15],
- Felzenszwalb person and car detector [16].

### 3.1.4. Central Priority

Humans generally place the object of interest near the center of the image while taking pictures. For this reason, a feature which specifies the distance to the center for each pixel is used.

### 3.1.5. SIFT Keypoints

For objects in a scene, the interesting points on the object can be extracted to provide a feature description of them. This description can be used to identify the object in a test image. The SIFT [17] method can be used to detect these interesting points in an image. In this paper, the keypoint localization step of the method is used. Low-contrast keypoints are discarded and locations of the more interesting keypoints, which can be attractive to the human eye, are used. One contribution of this paper is the using of the position of interest points on an image as a feature and increasing the classification rate. Other local features, like SURF and GLOH were also experimented with, but since SIFT had a better effect on the model, it was SIFT that we decided to go with.

### 3.2. Training Phase

While training and testing the saliency model, machine learning methods described in Section 2.2 were used. Images in the database were divided up as 80% training and 20% testing. That means that 803 images were used for training and 201 for testing. 10 positively labeled pixels from the most 5% salient locations and 10 negatively labeled pixels from the least 30% salient locations were chosen from each image. Thus the examples are guaranteed to be both strongly positive and strongly negative. Any samples on the boundary between the two regions, and samples within 10 pixels of the image boundary were not selected. It is noted that selecting more than 10 samples as positive and negative did not increase the classification rate and contained redundant information.

The dataset used to train the model consisted of 16,040 samples; the test dataset consisted of 4,020 samples. Every sample (point) has 34 continuous attribute and labeled as positive or negative. Obtained data was given as an input to the described learning methods for training and a more accurate model will be discussed. Details of the parameters used in the classifier training are explained below:

- In the model that uses SVM, all data were normalized and radial basis function (1) is used for kernel function (other kernels produce results similar to radial). γ value is assigned to 0.8. Cost of misclassification didn't affect the result, so *c* is assigned to 8.

$$k(x_i, x_j) = exp\left(-\gamma \|x_i - x_j\|^2\right) \qquad (1)$$

- In the model that uses C4.5 decision tree, for spanning variables, the minimum number of objects (instances) in leaves $m = 2$ and pruning confidence level (*cf*) is assigned to 25%.
- The number of nearest neighbors (*k*) used in classification is 9 for the kNN method and it was observed that the value does not affect the result. In this model all attributes are normalized and Euclidean distance is used as a distance metric.
- In the NaiveBayes method, we used relative frequency as a prior in probability estimation. Size of LOESS window is 0.5 and LOESS sample points are set to 100.
- Adaboost method is used with SVM to improve the result obtained. Boosting with 10 SVM classifier is applied and weighted combination of weak learners is obtained.

### 3.3. Comparison of Learning Methods

In this section classification performance of machine learning methods will be discussed. While sampling, a 5-fold cross validation is used and how the specified model accurately worked in practice is shown.

Table 1.  Evaluation results of the methods using same features

| Method | Results (same features) | | | | | |
|---|---|---|---|---|---|---|
| | *CA* | *Sens* | *Spec* | *AUC* | *Prec* | *Recall* |
| **SVM [1]** | 0,8801 | **0,8825** | 0,8778 | **0,9597** | 0,8786 | **0,8825** |
| **C4.5** | 0,8410 | 0,8413 | 0,8407 | 0,8730 | 0,8409 | 0,8413 |
| **kNN** | 0,8583 | 0,8407 | 0,8759 | 0,949 | 0,8695 | 0,8407 |
| **NaiveBayes** | 0,8168 | 0,8209 | 0,8129 | 0,9029 | 0,8147 | 0,8209 |
| **AdaBoost** | **0,8821** | 0,8818 | **0,8825** | 0,8818 | **0,8823** | 0,8818 |

Table 2.  Evaluation results of the methods using same features and SIFT

| Method | Results (same features+SIFT) | | | | | |
|---|---|---|---|---|---|---|
| | *CA* | *Sens* | *Spec* | *AUC* | *Prec* | *Recall* |
| **SVM** | 0.9065 | **0.9089** | 0.9091 | **0.9884** | 0.9091 | **0.9089** |
| **C4.5** | 0.8610 | 0.8656 | 0.8611 | 0.8971 | 0.8649 | 0.8659 |
| **kNN** | 0.8820 | 0.8629 | 0.9059 | 0.9791 | 0.8991 | 0.8665 |
| **NaiveBayes** | 0.8423 | 0.8405 | 0.8327 | 0.9282 | 0.8355 | 0.8482 |
| **AdaBoost** | **0.9085** | 0.9072 | **0.9089** | 0.9099 | **0.9087** | 0.9055 |

Table 1 and 2 shows the evaluation results of the methods used in this study. As a key: *CA* means classification accuracy, *Sens* means sensitivity, *Spec* means specificity, *AUC* means area under ROC curve, *Prec* means precision, *Recall* as itself.

Values found in the evaluation results show how the models can be compared to human performance while looking at an image. The study [1] that our model is based on used the SVM model (Table 1) and reached 88% classification accuracy. Similarly in our study, the SVM model gives better results than other methods. The main reason for this performance is that our problem is based on a binary classification prediction (positive and negative points). Also the dataset contains high-dimensional data with large numbers, and this makes SVM more advantageous over the other methods.

C4.5 decision tree method produced lower accuracy results because it doesn't have a suitable structure for large datasets. In addition, it falls behind on computation time during training when compared to other methods. In this work, Random Forest method was also used, but because of the failure of the decision trees on this dataset and because the algorithm couldn't construct more than 10 trees due to computational time limits, this method has not been reviewed in this paper.

The model which uses kNN method produced a lower accuracy because the error rate increases in multi-dimensional space with this method. In addition, computational time of the testing stage is very high when compared to other methods. Evaluation results show that the NaiveBayes method is the least successful. Especially as dependencies between attributes (for example, color features and their probabilities are dependent) leads this method to failure.

The Adaboost method produced similar results with SVM; even better in some evaluations. That result is obvious because it used the successful method SVM, focusing more on previously misclassified samples iteratively and constructed a better model. Saliency maps obtained from some sample images using this model are shown in Figure 6.

Another comparison between learning methods is given in Figure 1 using ROC curve.

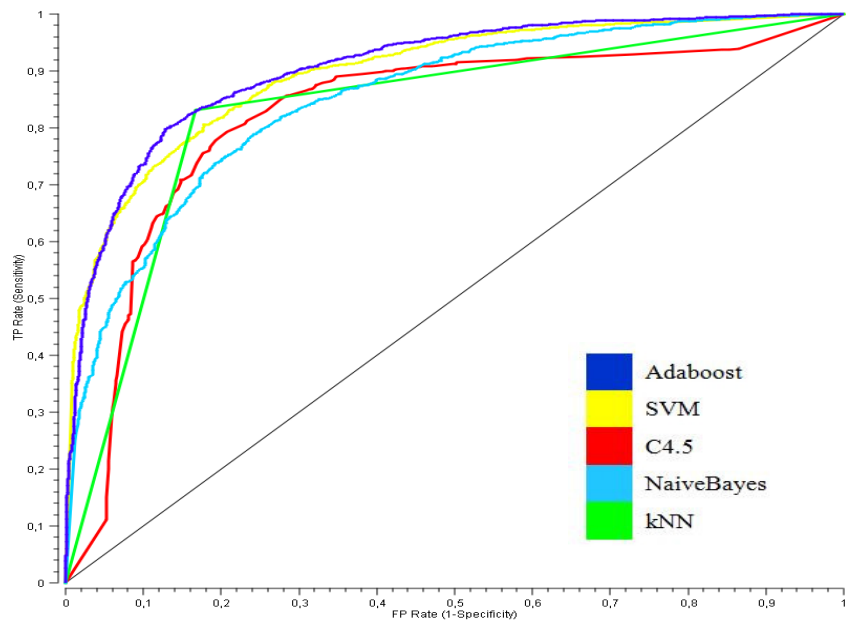

Figure 1.  ROC curves of methods (Adaboost, SVM, kNN, C4.5, NaiveBayes)

## 4. CONCLUSIONS

In this study, visual saliency models which try to predict where humans look at images are developed. For this purpose, a real eye-tracking database is used and different levels of features are extracted from the images. Using eye fixation points of 15 viewers, a ground truth saliency map (Figure 5) is uncovered with this information when different learning methods are trained and classification results are compared. As a conclusion of this study, machine learning methods that can be applied to this dataset are explained with reasons. Another contribution of this study is that the interest points extracted from the SIFT method are added to the feature set and better classification results (~90%) were then obtained over previous studies.

As a future piece of work, we expect to speed up the image processing step and feature extraction. However, training time of the model is not important because it can be done offline, but when a system wants to find the most salient points or regions on an image, this process must be handled quickly. If online saliency is successful, it can be integrated with many applications and can also be used in mobile systems.

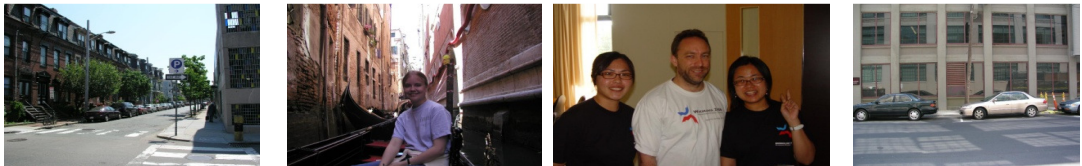

Figure 2.  Some example images from the database.

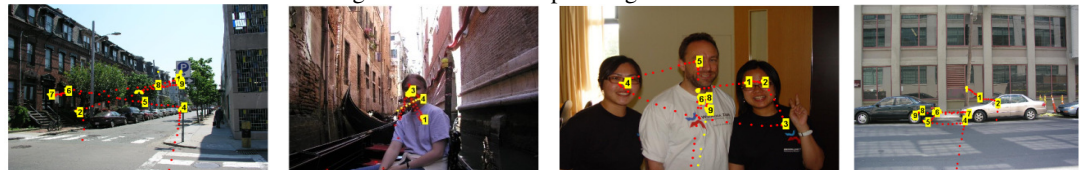

Figure 3.  Eye fixation points of one observer on the images.

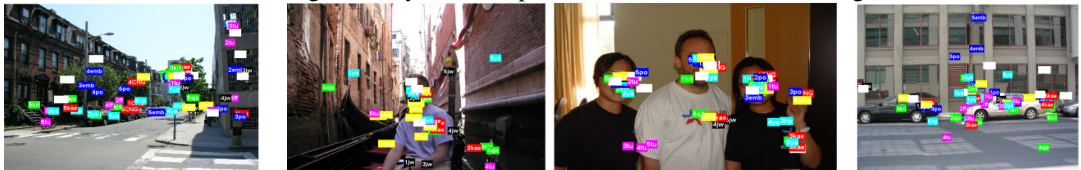

Figure 4.  Eye fixation points of all observers on the images.

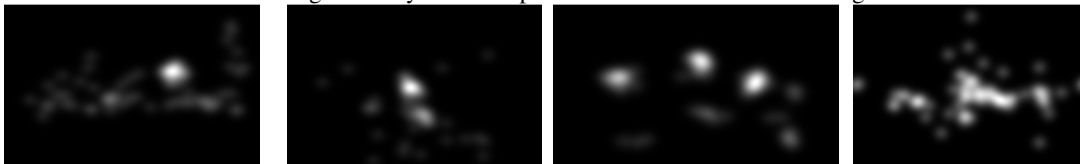

Figure 5. Continuous saliency map obtained from convolving Gaussian over fixation points of all observers.

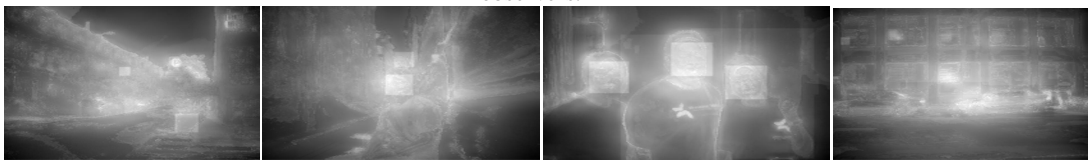

Figure 6.  Saliency map obtained from Adaboost

## AUTHOR

**H. Yalın Yalıç** was born in Ankara, Turkey in 1984. Yalıç received his B.Sc. degree from the Department of Computer Engineering, Çankaya University, Ankara, Turkey in 2007. He received his M.Sc. degree from the Department of Computer Engineering, Hacettepe University, Ankara, Turkey in 2010 and has continued with his Ph.D. education in the same department. He has worked as a Research and Teaching Assistant at Hacettepe University Computer Engineering Department since 2007. The author's major field of study is computer vision, image and video processing. 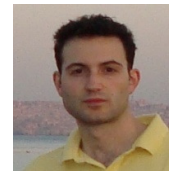